%% file: main.tex
\documentclass[11pt]{article}

\usepackage[preprint]{acl}

\usepackage{times}
\usepackage{latexsym}

\usepackage[T1]{fontenc}

\usepackage[utf8]{inputenc}

\usepackage{microtype}

\usepackage{inconsolata}

\usepackage{graphicx}
\usepackage{booktabs}

\usepackage{siunitx}
\usepackage{fancyhdr}

\usepackage{xcolor}
\definecolor{tokengray}{gray}{0.35}

\title{onPanda: Efficient Annotation of On-Policy Alignment Data for LLMs and Agents via Token-Level Correction}

 \author{
  \\
    \textbf{Lei Yang\textsuperscript{1}},
    \textbf{Mengyin Liu\textsuperscript{1,2}},
    \textbf{Jia Wang\textsuperscript{1}},
    \textbf{Hangyu Guo\textsuperscript{1}},
  \\
    \textbf{Liang Zhao\textsuperscript{1}},
    \textbf{Zheng Ge\textsuperscript{1}},
    \textbf{Kang An\textsuperscript{1}},
    \textbf{Binxing Jiao\textsuperscript{1}},
  \\
    \textbf{Qi Han\textsuperscript{1}},
    \textbf{Daxin Jiang\textsuperscript{1}},
    \textbf{Siqi Shen\textsuperscript{2}},
    \textbf{Xiangyu Zhang\textsuperscript{1}}
  \\[-1em]
  \\
    \textsuperscript{1}StepFun
    \qquad
    \textsuperscript{2}Xiamen University
  \\
  \\[-0.7em]
    \small{
      \href{https://on-panda.github.io/research}
           {https://on-panda.github.io/research}
           \qquad
           \texttt{yanglei@stepfun.com}
    }
  }

\begin{document}
\maketitle

\input{sec/0_abstract}

\input{sec/1_introduction}



\input{sec/2_onPanda}

\input{sec/3_evaluation}

\input{sec/4_benchmark}

\input{sec/6_conclusion}



\bibliography{reference}
\clearpage
\appendix

\input{sec/8_appendix}

\input{sec/5_related_work}

\input{sec/7_limitations_and_ethics}

\input{sec/9_future_work}

\end{document}

%% file: sec/0_abstract.tex
\begin{abstract}
We present onPanda, an interactive tool for efficiently annotating LLM alignment data and agent trajectories. onPanda adopts token-level correction as its core interaction: while reading a model response, the annotator locates the first inappropriate token and either picks a substitute from the model's candidate tokens or types the correct text via free-form editing. The system then truncates everything after that position and continues generation from the corrected prefix, repeating this locate-correct-continue loop until a satisfactory response is obtained. This mechanism lets annotators precisely steer model outputs at low cost: a small controlled study suggests that onPanda reduces median annotation time by 52\% over manual post-editing. Since the vast majority of tokens in the final response are generated by the model itself, the resulting data largely preserves the model's sampling distribution and is well suited for constructing on-policy SFT and preference data. Furthermore, the token-level corrections recorded during annotation provide fine-grained supervision with precise positions and naturally paired positive--negative samples. onPanda also connects to external tools and harnesses, enabling interactive trajectory annotation in realistic environments. In addition, we release Panda-CVL, a dataset annotated with onPanda, together with a benchmark for token-level correction. 
\end{abstract}

%% file: sec/1_introduction.tex
\begin{figure}[!t]
  \centering
  \includegraphics[width=\columnwidth]{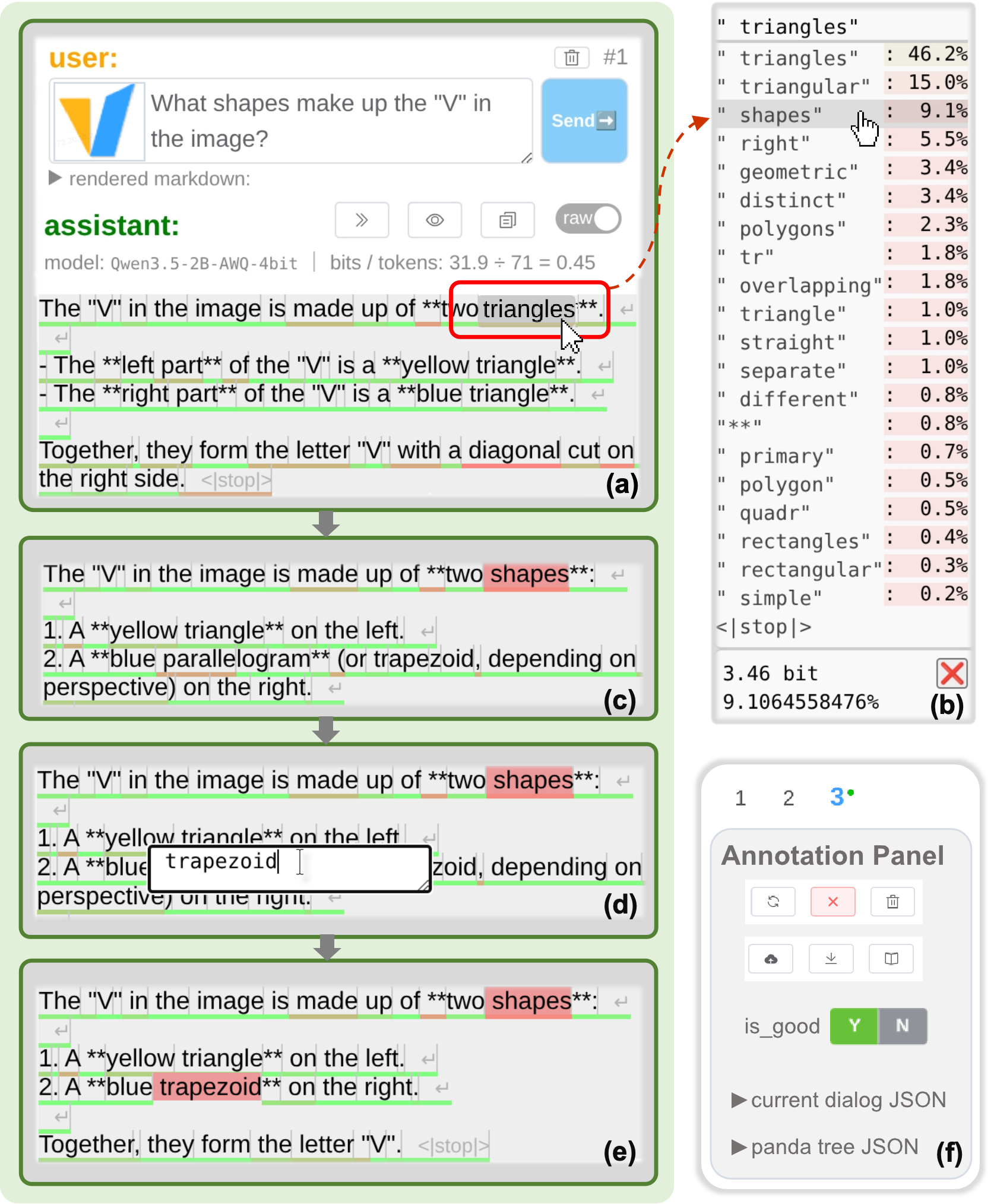}
  \caption{\textbf{The token-level correction interface of onPanda.}
  (a)~Each token's generation probability is color-coded below it: greener means higher, redder means lower.
  (b)~Hovering over the first inappropriate token \texttt{\textvisiblespace triangles} pops up the top-20 candidate tokens at this position; the annotator selects the more appropriate \texttt{\textvisiblespace shapes}.
  (c)~Replace and continue: the system replaces \texttt{\textvisiblespace triangles} with \texttt{\textvisiblespace shapes} (highlighted with a red background), truncates the subsequent content, and lets the model continue generating from the corrected prefix.
  (d)~Free-form editing: when no candidate fits, the annotator double-clicks the erroneous \texttt{\textvisiblespace paralle} to open an input box and replaces it with the correct \texttt{\textvisiblespace trapezoid}.
  (e)~The model continues generating after \texttt{\textvisiblespace trapezoid}, finally yielding a response that meets the SFT quality bar (\texttt{is\_good=Y}).
  (f)~The session preserves all intermediate versions: (a), (c), and (e) correspond to Dialogs [1, 2, 3]; the final Dialog~3 is the positive sample (\texttt{is\_good=Y}), while Dialogs~1 and~2 are retained as negatives.}
  \label{fig:ui}
  \vspace{-10pt}
\end{figure}

\section{Introduction}

Scaling high-quality data remains the central bottleneck for LLM alignment and agents. Existing pipelines struggle to reconcile annotation cost, on-policy fidelity, and supervision granularity: manually writing~\cite{ouyang2022training,nakano2022webgptbrowserassistedquestionansweringhuman} or post-editing~\cite{yao2023improving,scheurer2024traininglanguagemodelslanguage,schick2022peer} responses for SFT~\cite{ranzato2016sequence,arora2022exposure} is costly and yields off-policy data, while preference data is cheaper but offers only coarse response-level supervision~\cite{lightman2023lets,uesato2022solving,wu2023fine} and, being confined to candidates the model can already sample, provides little corrective signal when the model cannot sample good responses. Annotating agent trajectories is even harder: a trajectory arises from multi-step environment interaction, so any correction must actually execute tools and obtain real feedback before generation continues. Existing tools~\cite{jurgens2026potato,ou2025agentdiagnose} can observe and score live agents and intervene mid-run, and Reptile~\cite{reptile2025} further lets annotators edit model-output text that its terminal environment then executes; yet the community still lacks an interactive tool that, across diverse environments, both corrects general reasoning and tool calls in real time and executes the corrected calls. We therefore develop onPanda (\textbf{on}-\textbf{P}olicy \textbf{a}lig\textbf{n}ment \textbf{d}ata \textbf{a}nnotator), an interactive tool for efficiently annotating alignment data and agent trajectories.

onPanda's token-level correction interaction lets annotators precisely control and steer LLM generation, as illustrated in Figure~\ref{fig:ui}. While reading a model response, the annotator only needs to locate the first inappropriate token; onPanda displays the model's top-$k$ candidate tokens with their probabilities at that position: the annotator simply clicks a suitable substitute if one exists, or double-clicks the token and types arbitrary replacement text when the candidates miss the correct direction. After each correction, the system truncates the subsequent content and lets the model continue from the updated prefix---hence the annotator no longer needs to re-read the entire response after a fix, but simply repeats this ``locate--correct--continue'' loop along the response until it meets the requirements.

This paradigm directly addresses the above bottlenecks, bringing four advantages:
\textbf{(1)~Efficient annotation.} Most corrections are mouse clicks, with brief typing only when the candidates fall short---a far lower cognitive and typing burden than writing from scratch or manual post-editing; the linear ``correct-as-you-read'' workflow also avoids repeated review of the entire response, further cutting time cost.
\textbf{(2)~High on-policy fidelity.} Human intervention touches only the few erroneous positions; the vast majority of tokens are generated by the rollout model itself, either sampled directly from the prompt or continued from a corrected prefix, and corrections drawn from the model's own high-probability candidates perturb the sampling distribution even less. Unlike manual post-editing, the resulting SFT data thus stays close to the rollout model's distribution.
\textbf{(3)~Fine-grained supervision.} Alongside the SFT data, onPanda automatically records every token-level correction, capturing the error position, the substitute token, and a naturally paired positive--negative sample. These signals are finer-grained than response-level preferences: they can be converted into preference data for reward model or DPO training, or into process reward data for PRM training, based on the step containing the corrected token and its preceding steps.
\textbf{(4)~Complete expressiveness.} Free-form editing is a fallback beyond candidate selection, freeing annotation from the limits of the top-$k$ candidate set: even when the model cannot sample the correct content on its own---where preference annotation can hardly help---the annotator can still inject the correct text and steer subsequent generation. Such injections deviate from the model's own distribution, but occur only at the few positions beyond its capability, trading a minimal distributional cost for correctness. Hence, whenever the annotator can recognize and supply the correct fix, onPanda can construct SFT samples that meet the task requirements.

For agent settings, onPanda can access tools via MCP and connect to harnesses such as Claude Code, Codex, and OpenClaw, enabling interactive trajectory annotation in realistic environments. Its intuitive interface further suits model inspection.

In addition, we release Panda-CVL (a \textbf{C}hinese \textbf{V}ision--\textbf{L}anguage dataset for token-level correction), annotated with onPanda, together with an accompanying benchmark, to facilitate community research on this new type of data.

Our contributions are as follows:
\begin{itemize}
    \item We propose onPanda, an annotation tool centered on token-level correction for efficiently annotating LLM alignment data and agent trajectories that stay close to the rollout model's distribution.
    \item We experimentally validate onPanda's annotation efficiency and the on-policy fidelity of the produced data, and demonstrate its support for multimodal and agent-trajectory annotation.
    \item We release the multimodal Panda-CVL dataset with an accompanying benchmark, providing public resources for research on token-level correction data.
\end{itemize}

%% file: sec/2_onPanda.tex
\section{Design and Implementation}
\subsection{System Overview}

onPanda is a componentized front-end library with two usage modes. As a lightweight web app, it works out of the box: deployed via static hosting without any database, it lets users drag in local annotation files and start annotating, with the browser sending requests to preset or custom Chat Completions APIs. Alternatively, its components can be embedded into an existing data platform whose backend handles task dispatching and data collection---the form in which onPanda is integrated into our in-house production annotation system. onPanda only requires the inference API to (i)~continue generation from an assistant-message prefix (e.g., vLLM's \texttt{continue\_final\_message}) and (ii)~return each token's top-$k$ candidates with probabilities (logprobs). All artifacts of an annotation session---messages, annotations, operation logs, and probability caches---are serialized into a single \texttt{.panda.json} file for easy distribution, collection, and parsing.

\subsection{Token-Level Correction Engine}

Generation requests stream with logprobs: onPanda records each token's sampled probability and top-$k$ (default 20) candidates, and renders the corresponding probability color below each token (Figure~\ref{fig:ui}a). Since tokenizers may split a multi-byte character or emoji across tokens, the UI groups the token stream into minimal readable units (\emph{chunks}) along grapheme boundaries: interaction operates on chunks, while records stay token-precise. Each correction preserves three pieces of information: the correction position, the text replacement, and the samples before and after the correction.

After the annotator clicks a candidate or double-clicks to edit, onPanda truncates everything after the correction point and requests native continuation with the corrected response as the assistant prefix; tokens before the correction point are kept as-is rather than regenerated. Text introduced by free-form editing initially has no probability information; a single \texttt{prompt\_logprobs} request recomputes per-token probabilities and candidates for the entire response (this feature additionally requires \texttt{prompt\_logprobs} support from the API). This ``refresh'' also applies to arbitrary external text and cross-model settings: pasting any response into onPanda (or switching the model) reveals the current model's confidence on every token, making onPanda double as a model-inspection tool.
\vspace{-2pt}
\subsection{Annotation Tree and Data Protocol}

An annotation session is organized as an \emph{annotation tree} whose nodes are dialogs, each containing a complete copy of the messages and tools together with their annotations. Each correction forks a new node whose parent is the pre-correction dialog (Dialogs [1,2,3] in Figure~\ref{fig:ui} form a three-node chain: the initial rollout and two iterative corrections). All intermediate versions are thus saved automatically, without any version management by the annotator. Each node carries an operation log recording the operation type, timestamp, correction content, whether the operation is marked as on-policy, and a snapshot of the sampling configuration, giving every data point fully traceable provenance.

Each dialog also carries a quality verdict \texttt{is\_good} and a project-defined annotation schema (single-choice, multiple-choice, text, etc.). The companion Python library \texttt{onpanda} parses \texttt{.panda.json} into training data: nodes with \texttt{is\_good=Y} are exported as SFT samples, and positive--negative nodes under the same prompt are paired into response-level preference data. Within such pairs, the negative is typically an ancestor of the positive: the two share the prefix before the correction point and diverge exactly there, so the corresponding token-level correction data can be computed under any tokenizer as triples (negative sample, rejected token and its position, chosen token). Such data provides supervision that is precise in both position and update direction; positive and negative tokens pair one-to-one at the same position, so optimization receives naturally balanced signals. We believe these properties make token-level correction a promising cornerstone for more efficient post-training methods.

\vspace{-5pt}
\subsection{Agentic and Multimodal Annotation}

\begin{figure}[!t]
  \centering
  \includegraphics[width=\columnwidth]{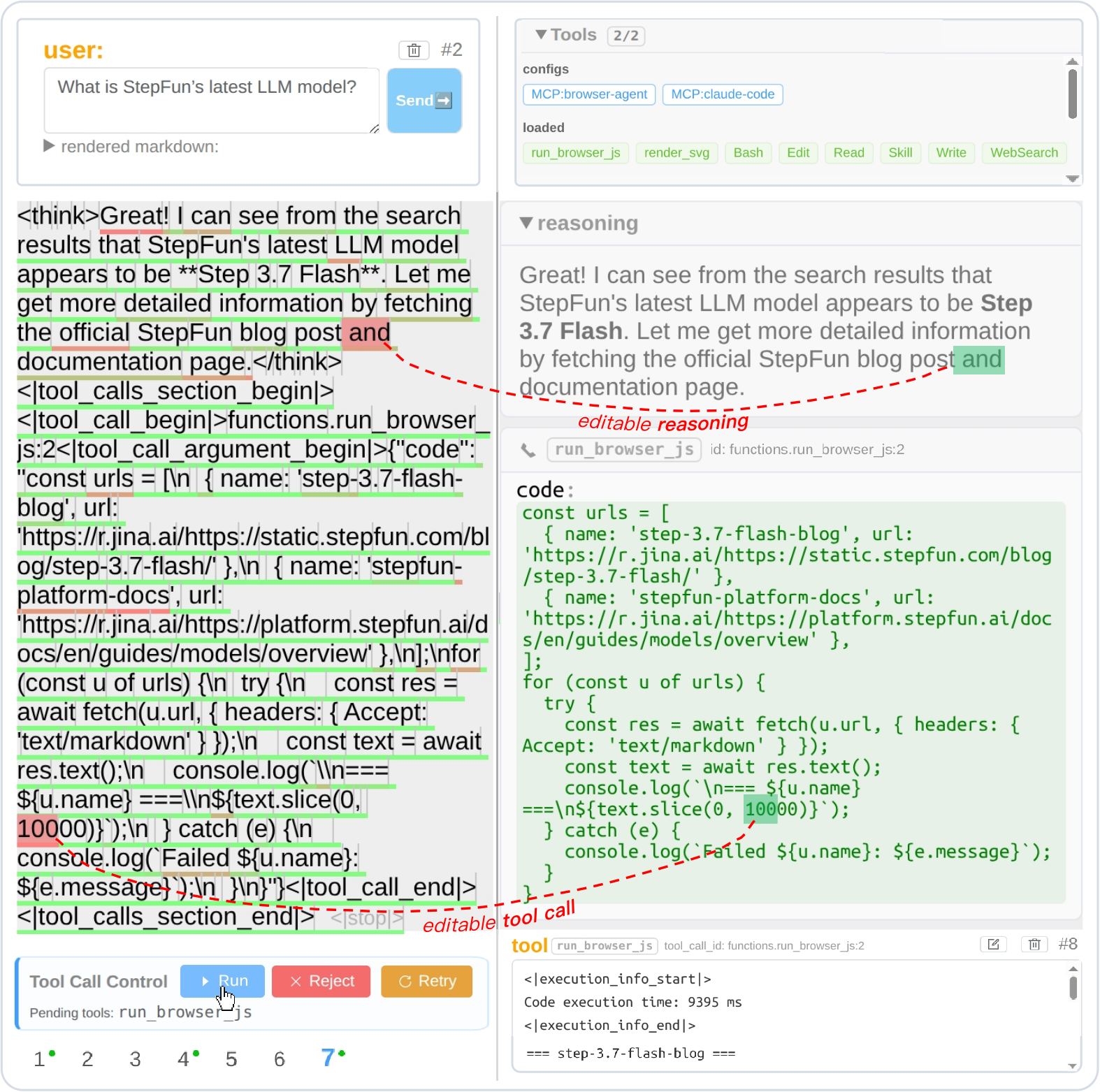}
  \caption{\textbf{Annotating an agent trajectory with onPanda.}
  Top left: the user prompt; top right: the connected MCP servers and tools.
  Left: the assistant response is rendered by the response template into the model's native token stream, where special tokens (e.g., \texttt{</think>}) are directly visible and correctable; the annotator corrects tokens in the reasoning and in the tool-call arguments, then approves execution.
  Right: the modified token stream is parsed in real time into \texttt{reasoning} and \texttt{tool\_calls}, displayed separately, with the returned tool results shown at the bottom right.}
  \label{fig:agent}
\end{figure}

Responses of reasoning models and agents are not plain text but structured messages containing \texttt{reasoning}, \texttt{content}, and \texttt{tool\_calls}, which resist direct token-level correction and continuation. onPanda addresses this with the \emph{response template} mechanism, which converts bidirectionally between structured messages and the model's native token stream: the rendering direction produces, per the model's response template, the full token sequence with special tokens (e.g., \texttt{</think>}, \texttt{<|tool\_call\_begin|>}) for display, correction, and continuation; the parsing direction restores the generated stream into structured messages in real time for storage and tool execution. Special tokens are directly visible and correctable by annotators, so token-level correction uniformly covers reasoning chains, content, and tool-call arguments (Figure~\ref{fig:agent}). Since dialogs are stored in structured form and a model's response template is applied only during rendering and correction, the same data can be further corrected and continued by different models.

External environments and tools connect via MCP; the \texttt{harness\_to\_mcp} adapter wraps existing harnesses such as Claude Code, Codex, and OpenClaw into MCP servers for onPanda. Tool calls can be configured to await annotator approval before execution: a problematic call can be executed after its arguments are corrected, or rejected with optional textual guidance, and rejected trajectories are automatically kept as negative samples. Tool results are fed back into the context and generation continues, enabling interactive trajectory annotation in realistic environments. onPanda likewise supports inputting, displaying, and annotating image, audio, and video messages.

%% file: sec/3_evaluation.tex
\section{Evaluation}


\subsection{Annotation Efficiency and Data Quality}

\paragraph{Baselines.}
We compare against two mainstream annotation paradigms: (1)~\emph{manual post-editing}: the answer box is pre-filled with the model's initial rollout, which the annotator edits until it meets the SFT quality bar; (2)~\emph{preference ranking}: four rollouts are pre-generated per prompt; the annotator ranks them and records whether the best one qualifies as SFT data. The two paradigms are instantiated with the widely used POTATO and Argilla, both full-featured, freely configurable platforms that we set up as the most typical alignment-data workflows. Because each platform differs in both its interface and annotation paradigm, this comparison evaluates complete workflows, making it difficult to determine whether the observed gains arise from the interface, the paradigm, or their combination.

\paragraph{Setup.}
Three annotators labeled 21 image-description prompts, evenly split into 3 groups. A Latin-square design rotates the order of the three methods: each group is annotated exactly once per method and no annotator labels the same prompt with different methods, balancing prompt difficulty, individual proficiency, and ordering effects. None of the annotators are onPanda developers, and all completed training and warm-up tasks on all three methods. Initial rollouts are generated by the same model, Qwen3.5-35B-A3B (instruct mode), with the officially recommended sampling parameters (temperature 0.7, top-$p$ 0.8); onPanda, POTATO, and Argilla's first candidate share the same rollout to reduce sampling randomness.

\paragraph{Metrics.}
We report five metrics. (1)~\emph{Time}: median annotation time, with the mean also reported; the median is robust to the long tail of difficult prompts, while the mean reflects total human cost. (2)~\emph{Pairwise win rate}: for each prompt, the three outputs---the final responses from onPanda and POTATO and Argilla's top-ranked response---are compared pairwise by GPT-5.5, with each pair evaluated in both orders to reduce position bias. (3)~\emph{PPL}: response perplexity under the rollout model, used to measure on-policy fidelity. We independently sample four rollouts per prompt, use their mean PPL as the baseline, and report both each method's PPL and its relative change ($\Delta$PPL). (4)~\emph{SFT coverage}: the proportion of prompts that yield a qualified SFT response; under preference ranking, all four rollouts may be unqualified. (5)~\emph{Preference pairs}: the average number of preference pairs obtained per prompt.

\paragraph{Results.}
As Table~\ref{tab:tool-comparison} shows, onPanda's median time is 330\,s per prompt, 51.5\% less than POTATO's 681\,s and on par with Argilla's 336\,s; in means, onPanda (515.6\,s) is 27.5\% and 24.7\% lower than POTATO (711.1\,s) and Argilla (684.5\,s). Argilla's mean far exceeds its median, consistent with the long right tail of reviewing four lengthy responses on hard prompts. Part of the gain over POTATO stems from continuation: a hallucination or error often recurs across a response---post-editing must spot and fix every occurrence, whereas onPanda corrects only its first occurrence and the continuation stays consistent with the fix. onPanda also attains the highest pairwise win rate (66.7\%), suggesting that the efficiency gain does not come at the cost of quality; an anonymized human comparison further supports this result: human evaluators preferred onPanda over POTATO in 54.8\% of pairs.

On on-policy fidelity, onPanda and Argilla reach PPL 1.181 ($\Delta$PPL $+0.86\%$) and 1.161 ($-0.83\%$), both within 1\% of the baseline (1.171) and within re-sampling noise (per-prompt PPL fluctuates by about $\pm 2.8\%$ across rollouts); POTATO yields 1.596 ($+36.31\%$). onPanda thus preserves the rollout model's sampling characteristics much better than manual post-editing. Meanwhile, Argilla selects a qualified SFT response on only 11/21 (52\%) prompts, whereas the editing-based POTATO and onPanda keep editing until the response qualifies, reaching 100\% coverage by design. onPanda further derives 7.43 preference pairs per prompt, precisely aligned to correction positions, versus 6 from Argilla's four-way ranking and 0.95 from POTATO's pre-/post-edit pair (one initial rollout qualified without edits).

\begin{table*}[t]
\centering
\small
\setlength{\tabcolsep}{3pt}
\begin{tabular}{llcccrrcc}
  \hline
  \textbf{Tool} &
  \textbf{Paradigm} &
  {\textbf{Time (s)}}$\downarrow$ &
  {\textbf{Win rate}}$\uparrow$ &
  {\textbf{PPL}} &
  {\textbf{$\Delta$ PPL}} &
  {\textbf{SFT Cov.}}$\uparrow$ &
  {\textbf{Pref. pairs}}$\uparrow$ &
  {\textbf{NASA-TLX}}$\downarrow$ \\
  \hline

  Argilla &
  rank 4 candidates &
  336 (684) &
  28.6\% &
  1.161 &
  - \textbf{0.83\%} &
  52\% (11) &
  6.00 &
  5.4 \\

  POTATO &
  post-editing &
  681 (711) &
  54.8\% &
  1.596 &
  + 36.31\% &
  \textbf{100\%} (21) &
  0.95 &
  6.8 \\

  \textbf{onPanda} &
  token-level correction &
  \textbf{330} (\textbf{516})  &
  \textbf{66.7\%} &
  1.181 &
  + 0.86\% &
  \textbf{100\%} (21) &
  \textbf{7.43} &
  \textbf{3.1} \\

  \hline
\end{tabular}

\caption{
Comparison of annotation paradigms in the controlled study.
Time is reported in seconds as the median annotation time over the
21 prompts, with the mean in parentheses.
Win rate is the fraction of LLM-judged pairwise comparisons each
method wins against the other two.
PPL is scored by the rollout model; $\Delta$PPL is the relative
change from the re-sampling baseline (1.171); smaller
absolute values indicate higher on-policy fidelity.
SFT Cov. is the fraction of prompts yielding a valid SFT response,
and Pref. pairs is the mean number of derivable preference pairs per
prompt.
NASA-TLX, from the user study, is on our adapted 0--10 scale
(lower is better).
}
\label{tab:tool-comparison}
\end{table*}

\subsection{User Study and Real-World Deployment}

The three annotators then filled out an adapted NASA-TLX questionnaire following~\cite{nakano2022webgptbrowserassistedquestionansweringhuman} for each tool, rating workload on 6 subscales from 0 to 10. We averaged the six ratings per annotator and then across annotators, with lower scores indicating lower workload. onPanda achieved the lowest score (3.1), compared with Argilla (5.4) and POTATO (6.8). One annotator remarked: ``onPanda gives immediate feedback---I can see my progress on each item; the ranking workflow instead forced me to hold a very long context in mind, which was mentally taxing.'' Another noted: ``The probability shading speeds up locating what to fix---low-probability tokens often mean the model is less confident and more error-prone, so checking them first makes annotation faster.'' 

Since deployment, onPanda has continuously produced three types of production data---vision, audio, and agentic---with 25{,}596, 105{,}143, and 1{,}257 annotation sessions respectively, automatically yielding about 388K token-level corrections (see Table~\ref{tab:deployment} in appendix). Across all qualified responses, 97.0\% of tokens are model-generated, 2.1\% are selected from candidates, and only 0.9\% are manually typed, showing sparse human intervention at production scale.

\subsection{Model Inspection}


onPanda  supports model diagnosis through token-probability visualization. Given a selected model and arbitrary text, it displays per-token generation probabilities using color coding and entropy statistics. This helps users diagnose whether a faulty rollout originates from the model or from a pipeline error, inspect error probabilities and candidate distributions at critical positions, and combine probability analysis with token-level correction to steer generation along alternative paths. Together, these features provide fine-grained feedback for model behavior analysis and data quality auditing.

%% file: sec/4_benchmark.tex
\section{Dataset and Benchmark}

\paragraph{Dataset.}
To support research on token-level correction data and its downstream applications, we construct \textbf{Panda-CVL}, a publicly releasable subset of production data annotated using onPanda. We apply two filtering criteria: (1) all images are either created in-house or obtained from openly licensed sources (see the Ethics section); and (2) the tasks do not require internal business knowledge and instead evaluate general-purpose capabilities such as visual perception, image description, and visual reasoning. Panda-CVL is a predominantly Chinese vision-language dataset comprising 7,491 annotation sessions stored in the \texttt{.panda.json} format, with 6,839 sessions in the training set and 652 in the test set. The auxiliary rollouts used during annotation were generated using \texttt{step-1o-turbo}, a 32B-parameter dense VLM.

\paragraph{Benchmark.}
A single token-level correction performed by an annotator can be decomposed into three subtasks: (1) determine whether the response meets the quality standard for acceptable SFT data, in which case the annotator marks \texttt{is\_good=Y} and submits it; otherwise, (2) identify the first inappropriate token in the response; and (3) correct it to an appropriate token, either by selecting from candidate tokens or by manually editing it. Following this decomposition, we evaluate models on the Panda-CVL test set by asking them to perform token-level correction in the same manner as human annotators. Given a user prompt, and a candidate response, the model must determine whether the response is acceptable and, if not, locate and correct its first error. The first-correction position and the human-provided replacement recorded during annotation serve as the evaluation reference.

To support this evaluation, we design a prompt template that requires the model to produce its correction in a find-and-replace format. If the response requires no modification, the model should output \texttt{<|split|><|is\_good|><|split|>}. Otherwise, it should output \texttt{<|split|>}\{\texttt{matched\_text}\}\texttt{<|split|>}
          \{\texttt{matched\_index}\}\texttt{<|split|>}
          \{\texttt{replacement\_text}\}\texttt{<|split|>}.
The \emph{matched text} is a short span beginning at the first inappropriate token and is used to locate the error. The \emph{match index} identifies which occurrence to replace when the span appears multiple times; it is 0 in most cases. The \emph{replacement text} is a short correct span that replaces the match and begins the corrected continuation.

For the correction illustrated in Figure~\ref{fig:ui}(a)--(c), one valid model output is:
\begin{quote}
\small
\texttt{<|split|>triangles**.\textbackslash n\textbackslash n<|split|>0<|split|>}\\
\texttt{shapes**:<|split|>}.
\end{quote}

\paragraph{Evaluation Protocol.}
We expand the response trajectories in the 652 test conversations into 2,126 evaluation instances, comprising 652 final acceptable responses labeled as good and 1,474 intermediate responses labeled as not good. Because the evaluated models use different tokenizers, we require replacement spans to cover multiple tokens. Using the Qwen3.6 tokenizer, we compute correction triples for two pairs: (1) the ground-truth negative response and the final acceptable response from the same conversation, and (2) the ground-truth negative response and the response obtained by applying the model-generated correction. A prediction is considered correct if the two resulting triples are identical. This procedure enables tokenizer-agnostic evaluation across models. The complete prompt template and evaluation implementation are provided in the accompanying \href{https://github.com/on-panda/on-panda-python}{onpanda} repository.

\paragraph{Metrics.}
Table~\ref{tab:bench} reports the following metrics:
\textbf{Format:}
The proportion of outputs that can be successfully parsed and executed, measuring the model's instruction-following ability under the proposed template.
\textbf{GoodAcc:}
The proportion of ground-truth good instances for which the model correctly
outputs \texttt{<|is\_good|>}. This metric measures the model's ability to
avoid unnecessarily modifying already acceptable responses.
\textbf{Loc.-NG:}
The proportion of ground-truth not-good instances for which the model
correctly locates the ground-truth first-error position.
\textbf{Corr.-NG:}
The proportion of ground-truth not-good instances for which both the located
position and the replacement agree with the ground truth. This metric measures
end-to-end correction performance.
\textbf{F1:} The harmonic mean of GoodAcc and Corr.-NG,
used as the overall metric:
\[
\mathrm{F1}
=
\frac{2 \cdot \mathrm{GoodAcc} \cdot \mathrm{Corr.\text{-}NG}}
{\mathrm{GoodAcc} + \mathrm{Corr.\text{-}NG}}.
\]
This aggregation follows ProcessBench~\cite{zheng-etal-2025-processbench}.
Two trivial strategies---always predicting \texttt{is\_good} and always
attempting a correction---each yield a score of zero on one of the two
component metrics and therefore obtain an F1 score of zero.

\paragraph{Results.}
We evaluate recent multimodal models on the Panda-CVL test set. The complete results and analysis are provided in Appendix A (Table~\ref{tab:bench}). Our multi-reference analysis finds human location agreement far above chance, alongside measurable variation in correction positions and replacement tokens (Appendix~\ref{app:annotation-agreement}).

%% file: sec/6_conclusion.tex
\section{Conclusion}

We presented onPanda, an annotation tool centered on token-level correction. Through the locate--correct--continue loop, annotators precisely steer model generation with a few clicks and edits, efficiently producing alignment data and agent trajectories that stay close to the rollout model's distribution, while automatically retaining every correction as fine-grained supervision. Experiments and a user study show that onPanda reduces annotation time and workload while preserving data quality. We open-source onPanda and release the Panda-CVL dataset and benchmark, hoping the community will exploit token-level correction as a novel supervision signal for more efficient post-training.

\section*{Ethics Statement}

Participants in our efficiency experiment and user study are full-time data annotators at our institution; the tasks fall within their regular salaried duties, and participation was voluntary and informed. Survey results are reported only in aggregate, and no personally identifiable information is disclosed in this paper. The annotation tasks involve image-description content without offensive or high-risk material. The released Panda-CVL contains only images created by us or under open licenses, manually reviewed before release to exclude personally identifiable information and inappropriate content, and is intended for research use only. As a general-purpose annotation tool, onPanda yields data whose properties depend on its users' annotation guidelines; we encourage practitioners to follow responsible data curation practices.

%% file: sec/8_appendix.tex
\section{ Additional Results and Statistics }
\label{sec:appendix}

\paragraph{Detailed Results.}
Table~\ref{tab:bench} presents the full evaluation results on the Panda-CVL test set.
Token-level correction remains challenging for all evaluated models, with the best F1 reaching only 17.09\%.
GPT-5.5 achieves the highest GoodAcc (53.37\%) and the best overall F1 (17.09\%),
whereas GPT-6 obtains the highest Loc.-NG (24.46\%) and Corr.-NG (15.83\%).
GPT-5.6-sol achieves the best Format score (99.98\%) but performs less strongly on GoodAcc.
Seven of the nine reasoning models achieve Format scores above 90\%, while Corr.-NG remains below 16\% for all of them,
suggesting that reliable format compliance does not necessarily translate into accurate error localization and correction.
The substantial variation in GoodAcc further shows that models differ considerably in their tendency to accept or modify candidate responses.

\begin{table}[t]
  \centering
  \scriptsize
  \setlength{\tabcolsep}{2pt}
  \renewcommand{\arraystretch}{0.96}

  \begin{tabular}{
    @{}
    p{0.49\columnwidth}
    *{3}{>{\centering\arraybackslash}p{0.15\columnwidth}}
    @{}
  }
    \toprule
    \textbf{Metric}
      & \textbf{Vision}
      & \textbf{Audio}
      & \textbf{Agentic} \\
    \midrule

    \multicolumn{4}{@{}l}{\textbf{\textit{Scale}}} \\
    \addlinespace[2.5pt]
    ~~~~Annotation sessions
      & 25{,}596
      & 105{,}143
      & 1{,}257 \\
    ~~~~Annotators
      & 32
      & 42
      & 24 \\
    ~~~~Total human-hours
      & 3{,}557.4
      & 18{,}871.3
      & 523.5 \\
    ~~~~Annotation time P25 (min)$^\dagger$
      & 0.64
      & 0.95
      & 17.82 \\
    ~~~~Annotation time P50 (min)$^\dagger$
      & 1.65
      & 1.93
      & 31.31 \\
    ~~~~Annotation time P75 (min)$^\dagger$
      & 4.48
      & 3.81
      & 50.01 \\

    \cmidrule{1-4}
    \multicolumn{4}{@{}l}{\textbf{\textit{Data yield}}} \\
    \addlinespace[2.5pt]
    ~~~~SFT samples (\texttt{is\_good=Y})$^\dagger$
      & 1.001
      & 1.318
      & 6.018 \\
    ~~~~Token-level corrections$^\dagger$
      & 2.256
      & 3.034
      & 8.566 \\
    ~~~~Candidate clicks$^\dagger$
      & 1.684
      & 2.655
      & 5.410 \\
    ~~~~Double-click edits$^\dagger$
      & 0.572
      & 0.380
      & 3.156 \\
    ~~~~Regenerations$^\dagger$
      & 0.003
      & 0.544
      & 0.303 \\

      \cmidrule{1-4}
      \multicolumn{4}{@{}l}{
        \textbf{\textit{Per \texttt{is\_good} response (avg.)}}
        } \\
      \addlinespace[2.5pt]
    ~~~~Reasoning corrections
      & --
      & --
      & 0.34 \\
    ~~~~Content corrections
      & 2.24
      & 2.05
      & 0.05 \\
    ~~~~Tool-call corrections
      & --
      & --
      & 0.66 \\
    ~~~~Zero-correction rate (\%)
      & 49.1
      & 23.9
      & 53.5 \\
    ~~~~Model-generated tokens
      & 207.0
      & 34.7
      & 557.0 \\
    ~~~~Candidate-selected tokens
      & 1.6
      & 1.6
      & 0.9 \\
    ~~~~Annotator-typed tokens
      & 1.0
      & 0.7
      & 1.2 \\
      
    \cmidrule{1-4}
    \multicolumn{4}{@{}l}{\textbf{\textit{Task profile}}} \\
    \addlinespace[2.5pt]
    ~~~~Image inputs$^\dagger$
      & 1.026
      & --
      & 52.010 \\
    ~~~~Audio inputs$^\dagger$
      & --
      & 1.000
      & -- \\
    ~~~~Tool calls$^\dagger$
      & 0
      & 0
      & 5.230 \\
    ~~~~User turns$^\dagger$
      & 1.000
      & 16.163
      & 1.387 \\

    \bottomrule
  \end{tabular}

  \caption{ Production annotation statistics for the Vision, Audio, and Agentic onPanda deployments. $^\dagger$ denotes statistics reported per annotation session. Some projects require annotating multiple SFT instances from a single session to increase data throughput. Furthermore, a single agent trajectory contains multi-turn tool calls that can serve as SFT data. Due to the nascent state of agentic annotation and unoptimized rollout models, stricter control is required, leading to increased manual typing. } 
  \label{tab:deployment}
\end{table}

\begin{table}[t]
  \centering
  \scriptsize
  \setlength{\tabcolsep}{3pt}
  \renewcommand{\arraystretch}{0.96}

  \resizebox{\columnwidth}{!}{%
    \begin{tabular}{@{}lrrrrr@{}}
      \toprule
      \textbf{Model}
        & \textbf{Format}
        & \textbf{GoodAcc}
        & \textbf{Loc.-NG}
        & \textbf{Corr.-NG}
        & \textbf{F1} \\
      \midrule
        
      \multicolumn{6}{@{}l}{\textit{Reasoning}} \\
      Doubao-Seed-2.1-pro\cite{doubao_seed21}
        & 99.01
        & 9.66
        & 21.64
        & 13.70
        & 11.33 \\
      GPT-5.5 (xhigh)\cite{gpt55}
        & 93.96
        & \textbf{53.37}
        & 15.26
        & 10.18
        & \textbf{17.09} \\
      GPT-5.6-sol (xhigh)\cite{gpt56sol}
        & \textbf{99.98}
        & 16.56
        & 19.95
        & 13.09
        & 14.63 \\
    GPT-6 (xhigh)\cite{gpt6}
          & 99.65
          & 17.38
          & \textbf{24.46}
          & \textbf{15.83}
          & 16.57 \\
      Kimi-K2.6\cite{kimi26}
        & 92.41
        & 26.40
        & 18.47
        & 10.59
        & 15.11 \\
      Step-3.7-Flash\cite{step37flash}
        & 78.83
        & 40.09
        & 5.81
        & 3.32
        & 6.13 \\
     Step-5-Preview\cite{step5}
      & 95.63
      & 32.21
      & 16.42
      & 9.77
      & 14.99 \\
      Qwen3.6-35B-A3B\cite{qwen36_35b_a3b}
        & 68.58
        & 26.23
        & 9.57
        & 5.70
        & 9.36 \\
      Qwen3.5-397B-A17B\cite{qwen35blog}
        & 93.77
        & 20.40
        & 17.37
        & 10.99
        & 14.28 \\
        
      \midrule
      \multicolumn{6}{@{}l}{\textit{Instruct}} \\
      Qwen3.5-397B-A17B\cite{qwen35blog}
        & 38.83
        & 4.45
        & 1.63
        & 0.81
        & 1.38 \\
      \bottomrule
    \end{tabular}%
  }
  \caption{
    Scores on the Panda-CVL test set (\%).
    The best result in each column is shown in bold.
  }
  \label{tab:bench}
\end{table}

\section{Annotation Agreement}
\label{app:annotation-agreement}

\begin{figure*}[t]
  \centering
  \includegraphics[width=0.92\textwidth]{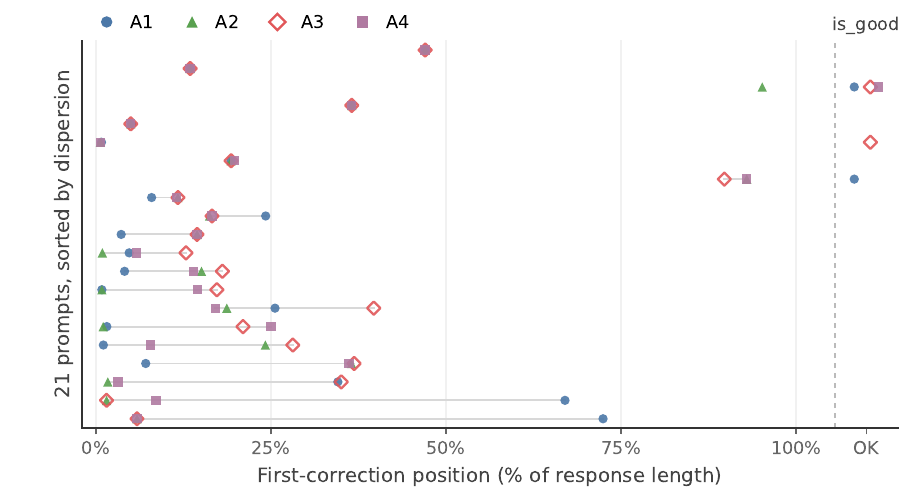}\par
  {\small (a)\par}
  \vspace{0.75em}
  \includegraphics[width=0.92\textwidth]{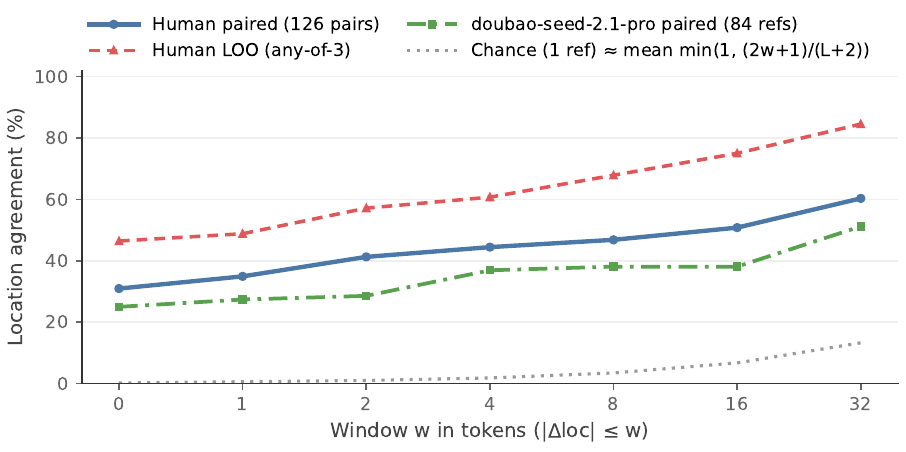}\par
  {\small (b)\par}
  \caption{Annotation agreement on Panda-MultiRef-21.
  (a)~Each row shows four independent annotations (A1--A4), ordered by correction-position spread; positions are normalized by response length $L$, and OK denotes no correction.
  (b)~Location agreement within $w$ tokens: human pairs (126 comparisons), human leave-one-out (matching any of the other three references), and Doubao-Seed-2.1-pro against its paired reference (84 comparisons).
  Positions are zero-based, with stop at $L$ and \texttt{is\_good} at $L+1$.
  Chance denotes the single-reference uniform random-location baseline, approximated by the prompt-wise mean of $\min(1,(2w+1)/(L+2))$.}
  \label{fig:multiref-iaa}
\end{figure*}

We analyze annotation agreement on \textbf{Panda-MultiRef-21}\footnote{\url{https://github.com/on-panda/Panda-MultiRef-21}}, comprising 21 image-description prompts with four independent onPanda annotations of the same initial response per prompt (84 annotations). We extract the first differing token between each initial and final accepted response, recording its position and replacement. Human pairwise location agreement is 30.95\% under exact matching, far above the uniform random-location baseline of 0.20\%; with a tolerance of four tokens, the rates are 44.44\% and approximately 1.84\%, respectively (Figure~\ref{fig:multiref-iaa}). This gap indicates shared localization judgments, supporting the reliability of the reference annotations. Meanwhile, agreement remains incomplete, and replacement-token agreement is 69.44\% conditional on identical correction positions. These observations characterize both consensus and diversity in the annotation distribution, providing context for interpreting single-reference exact-match scores.

%% file: sec/5_related_work.tex
\section{Related Work}

\paragraph{Annotation tools for LLMs and agents.}
General-purpose platforms such as POTATO~\cite{jurgens2026potato} and Argilla~\cite{argilla} can be configured for post-editing or preference annotation, and POTATO~2.0 further supports online and offline agent-trajectory annotation. Closest to our work is Reptile~\cite{reptile2025}: targeting terminal-based SWE tasks, it lets annotators locally edit model-output text to trigger continuation, which its terminal environment parses and executes before resuming the rollout, thereby yielding SFT data with higher on-policy fidelity than full manual post-editing. However, Reptile's corrections rely almost entirely on text manually typed by annotators rather than selected from the model's own candidate tokens, inevitably limiting the data's on-policy fidelity; its current implementation also encodes reasoning and actions as plain text and shell commands, and cannot support general tool-call formats. By contrast, onPanda combines selecting from probability-scored candidate tokens with continuation over structured reasoning and tool calls, covering plain-text LLM, multimodal, and agent data, and explicitly records each correction's position, replacement text, and pre-/post-correction samples. We do not compare against Reptile directly because it targets terminal-based SWE tasks with different environments and action representations; we compare workflows rather than benchmark the systems head-to-head on a shared task.

\paragraph{Fine-grained supervision signals.}
For fine-grained supervision, PRM800K~\cite{lightman2023lets} collects step-level labels through human annotation, while TLDR~\cite{tldr} constructs token-level labels from synthetic perturbations. In contrast, onPanda's token-level correction signals are position-precise and naturally paired as positives and negatives, arising directly from the annotator's process of constructing qualified responses, without extra human labeling or synthetic perturbation.

\paragraph{Correct-and-continue interaction.}
The pattern of ``correcting a prefix and regenerating the suffix'' predates the LLM era: Predictive Translation Memory~\cite{ptm} and INMT~\cite{inmt} already applied it to interactive machine translation. onPanda brings this mechanism into general-purpose alignment-data annotation and extends it to multimodal and structured agent scenarios, exporting branches of the annotation tree as three data types: SFT, preference, and token-level correction.

%% file: sec/7_limitations_and_ethics.tex
\section*{Limitations}

 The core interaction of onPanda relies on two inference-API capabilities: continuing generation from an assistant-message prefix, and returning top-$k$ candidate logprobs (probability refresh further requires \texttt{prompt\_logprobs}). Some proprietary APIs do not expose both, so onPanda cannot annotate on-policy data for the models behind them. In practice this constraint is mild: official APIs from many vendors---including the proprietary Doubao, as well as DeepSeek, Kimi, and StepFun---already provide both prefix continuation and logprobs; for open-weight models, mainstream inference frameworks such as vLLM and SGLang natively support all required features, llama.cpp and ollama cover the core interaction, and vLLM-based hosted services such as W\&B Inference\footnote{\url{https://wandb.ai/inference}} work with onPanda out of the box. Moreover, reasoning and tool-call continuation for a new model requires adapting its response template; with our reference implementations and development guide, a coding agent can usually complete the adaptation automatically, though it remains a one-time integration cost.

Second, the efficiency and on-policy advantages of token-level correction rest on a sparse-error premise: the model should produce responses whose local logic is largely sound, so that human intervention is needed only at a few positions. When the model falls far short of the target task and corrections become dense, both advantages diminish; annotators can then fall back to free-form editing, which onPanda also supports. Moreover, the on-policy property is relative to the rollout model at annotation time: its benefit fades when the data is used to train a different model, or as the trained model's weights are continually updated.

Finally, our empirical evidence has its own boundaries: the controlled experiment is modest in scale (3 annotators $\times$ 21 image-description prompts, a single rollout model), and the user-study participants all come from our in-house annotation team; quality evaluation primarily relies on LLM-as-a-judge, supplemented by a small human comparison between onPanda and POTATO; automated judgments may still reflect judge-specific preferences despite position-swapped comparisons. We do not conduct a controlled agent study; the agentic evidence is limited to system capability and deployment statistics. More importantly, this paper validates annotation efficiency, output quality, and the distributional properties of the produced data, whereas the downstream training benefit of token-level correction signals remains unverified by training experiments.

%% file: sec/9_future_work.tex
\section{Future Work}
\label{app:future-work}

\paragraph{Post-training on token-level correction data.}
Token-level correction data offers supervision that is precise in both position and direction, with naturally paired positive--negative signals. To verify its downstream training benefit, we are developing post-training methods built on such data: the \emph{token-level correcting model}, a token-level counterpart of reward models, and \emph{token-level correction optimization}, a token-level counterpart of DPO. The benchmark released with this paper evaluates exactly this correction capability, providing a ready testbed for the correcting model.

\paragraph{Annotation--training flywheel.}
A correcting model can in turn assist annotation by pre-locating suspicious tokens and proposing fixes for annotators to review. Moreover, onPanda can be integrated with a training pipeline into a closed loop: each newly annotated batch triggers an incremental training step and updates the served rollout model, which then assists subsequent annotation. As the model improves, corrections grow sparser and annotation keeps getting cheaper, while the data stays on-policy with respect to the latest weights---mitigating the on-policy decay noted in Limitations---so that annotation and training accelerate each other.

\paragraph{Online learning from textual feedback.}
onPanda's upcoming support for annotating earlier turns will yield data of a new form: guided by feedback that arrives in later turns (e.g., tool errors or user complaints), annotators revise an earlier rollout---future feedback supervising past generation. This mirrors the supervision structure of online learning from textual feedback, where deployed agents continually receive compiler errors, tool failures, and users' textual corrections, all pointing to concrete mistakes in earlier outputs. Historical-turn annotations could therefore become a natural data source for this direction: models (e.g., the correcting model) trained to translate such feedback into token-level corrections of earlier rollouts would let LLMs learn continually from real interactions without scalar rewards. In this vision, token-level correction serves as a unified intermediate representation that turns arbitrary textual feedback---from annotators, environments, or end users---into trainable supervision.